\documentclass{article}

\usepackage{microtype}
\usepackage{graphicx}
\graphicspath{{images/}}
\usepackage{subfigure}
\usepackage{booktabs} 
\usepackage{amsthm}
\usepackage{amsmath}
\usepackage{amsfonts}

\usepackage{hyperref}

\usepackage[accepted]{icml2019}

\icmltitlerunning{DetTransNet}

\begin{document}

\twocolumn[
\icmltitle{Visual Transformer for Object Detection}



\icmlsetsymbol{equal}{*}

\begin{icmlauthorlist}
\icmlauthor{Michael Yang}{oppo}
\end{icmlauthorlist}

\icmlcorrespondingauthor{Michael Yang}{xuewen.yang@protonmail.com.}

\icmlaffiliation{oppo}{OPPO Inc.}

\icmlkeywords{Machine Learning, ICML}

\vskip 0.3in
]



\printAffiliationsAndNotice{}  

\begin{abstract}

Convolutional Neural networks (CNN) have been the first choice of paradigm in many computer vision applications. 
The convolution operation however has a significant weakness which is it only operates on a local neighborhood of pixels, thus it misses global information of the surrounding neighbors.
Transformers, or Self-attention networks to be more specific, on the other hand, have emerged as a recent advance to capture long range interactions of the input, but they have mostly been applied to sequence modeling tasks such as Neural Machine Translation, Image captioning and other Natural Language Processing tasks. 
Transformers has been applied to natural language related tasks and achieved promising results.
However, its applications in visual related tasks are far from being satisfying.
Taking into consideration of both the weaknesses of Convolutional Neural Networks and those of the Transformers, in this paper, we consider the use of self-attention for discriminative visual tasks, object detection, as an alternative to convolutions.
In this paper, we propose our model: DetTransNet.
Extensive experiments show that our model leads to consistent improvements in object detection on COCO across many different models and scales, including ResNets, while keeping the number of parameters similar. 
In particular, our method achieves a 1.2\% Average Precision  improvement on COCO object detection task over other baseline models. 
\end{abstract}
\section{Introduction}

Convolutional Neural Networks have been having tremendous success in many computer vision and natural language processing applications, such as image classification \cite{Simonyan15}, object detection \cite{yolo}, recommendation \cite{xuewen_mm20}, image captioning \cite{Yang2020Fashion, Lu2017Adaptive,yang2019,Yao2018}, acoustic event detection \cite{feng2015}, neural SDE \cite{yingru}, and image to image translation \cite{yang2018cross}. 
Convolutional neural networks use local receptive field to focus on the local neighborhood of the pixels.
It also achieves translation equivariance through weight sharing.
These two properties are very important when design models to process images.
The weaknesses of CNN \cite{Kaiming16} is that convolutional kernels is not effective in locals because it prevents it from capturing global contexts in an image.
This limits the recognition performance because to achieve better performance, it is often necessary to be able to capture global contexts of the images.

Transformers \cite{Vaswani2017,yingruliu2,yang2019latent}, or self-attention, have overly replaced CNN in natural language processing tasks. They can capture long range interactions which proved to be an effective way of modeling. However, transformers have mostly been applied to sequence modeling tasks. 
Self attention is effective because it used a weighted average operation of the token features.
The weights are based on the similarity function between the features. It shows the interaction between the inputs.
This mechanism allows Transformer to capture long range interactions of sequences.
The positional encoding determines the location of the input tokens which is of vital importance.
Different from the pooling or the convolutional operator, the weights used in the weighted average operation are produced through learning via a similarity function between hidden units. 

As the first Transformer model in vision tasks, ViT (vision transformer \cite{dosovitskiy2021an}) proves the full-transformer architecture is promising for vision tasks. However, its performance is still inferior to that of similar-size CNN counterparts, for example, Faster R-CNN, when trained from scratch on a midsize dataset.
Through analysis, there are two main limitations of ViT. Firstly, the straightforward tokenization of input images by hard split makes ViT unable to model the image local structure like edges and lines, and thus it requires significantly more training samples than CNNs for achieving similar performance. Secondly, the attention backbone of ViT is not well designed as CNNs for vision tasks, which contains redundancy and leads to limited feature richness and difficulties in model training.

Transformers are a deep learning architecture that has gained popularity in recent years. They rely on a simple yet powerful mechanism called attention, which enables deep learning models to selectively focus on certain parts of their input and thus reason more effectively. Transformers have been widely applied on problems with sequential data, in particular in natural language processing (NLP) tasks such as language modeling and machine translation \cite{yang2019latent}, and have also been extended to tasks as diverse as speech recognition, symbolic mathematics, and 
reinforcement learning. But, perhaps surprisingly, computer vision has not yet been swept up by the Transformer revolution.

To help bridge this gap, we are releasing Detection Transformer Network (DetTransNet), an important new approach to object detection. DetTransNet completely changes the architecture compared with previous object detection systems. 
The contributions of this paper are:
\begin{itemize}
\item We introduce a novel concept of overlapping image patches to enable more accurate and flexible visual sequences.
\item We propose a Vision Transformer based object detector that takes an image as the input and outputs the objects classes as well as the bounding boxes. The experimental results demonstrates the effectiveness of our model.
\end{itemize}

The remainder of this paper is organized as follows.  Section~\ref{sec:related} reviews the research for object detection problems. Section~\ref{sec:method} describes our model and training method in details. Section~\ref{sec:exp} presents our evaluation metrics, experimental methodology and the evaluation results on the model accuracy and efficiency.
Finally, we conclude our work in Section~\ref{sec:con}.
\section{Related Work}
\label{sec:related}
In this section, we review most of the research work related to object detection.
\subsection{CNN Based Object Detectors}
The CNN based object detectors are mainly about R-CNN series, including R-CNN \cite{rcnn}, Fast R-CNN \cite{fast-rcnn} and Faster R-CNN \cite{Ren15}.
The basic idea behind the Region based CNN approach, such as R-CNN is to use bounding box detection as the objective and attend to some object region candidates. 
Then the final result is based on the object regions.
R-CNN was then improved by \cite{fast-rcnn} to better deal with the feature maps using ROIPooling, which results in faster speed and better accuracy.
Faster R-CNN \cite{Ren15} further improved this by learning an attention mechanism by using a Region proposal network. It is very efficient in many other papers that follow R-CNN series.

\subsection{Transformer Based Object Detectors}

\begin{figure}
    \centering
    \includegraphics[width=0.48\textwidth]{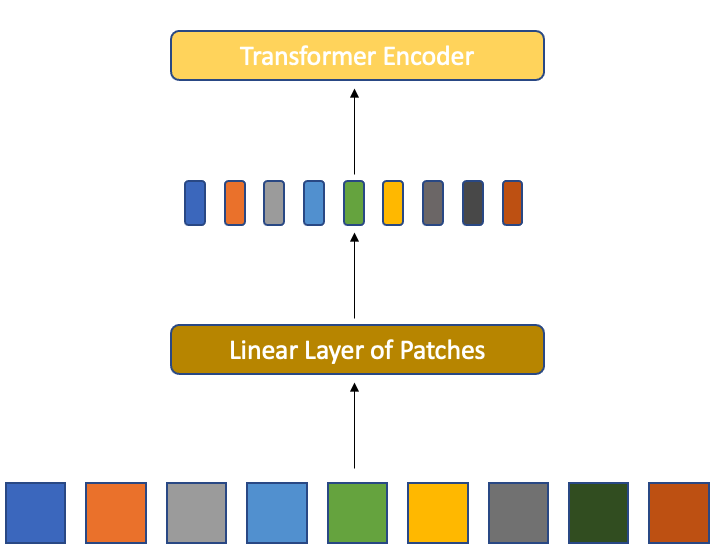}
    \caption{The Vision Transformer backbone.}
    \label{fig:vit}
\end{figure}

Vision Transformer or ViT \cite{dosovitskiy2021an} is the first Transformer based vision model. It simply treats the image as a sequence of patches and uses the original Transformer encoder to process image data. 
However, it can only generate a feature map of a single scale. Thus, it cannot detect small objects which should use objects of various scales (small-, mid-, or large-sizes). Usually, large objects can be easily detected at a rough scale of the image, while small objects are often detected at a finer scale. Several papers proposed to modify the architecture of ViT to generate multi-scale feature maps and demonstrated their effectiveness in object detection and segmentation tasks such as \cite{liu2021Swin, Wang_2021_ICCV}. The model architecture of ViT is shown in Figure \ref{fig:vit}.

\section{Methods}
\label{sec:method}
In this section, we present our model and how it solves the problem of object detection.
Our model can be easily applied to various Transformer architectures developed in the deep learning research fields, including computer vision and natural language processing. We design this characteristics not for competing with others in detection performance but for method development.

\subsection{High-Level Picture of the Model}
Our DetTransNet takes an image as the input. Then it divides the image into $n$ overlapping patches with $m$ pixels overlapped.
After that, the patches are embedded by a linear layer whose outputs are fed into $N$ Transformer encoder layers.
Then all $n$ overlapping patches are re-organized into an image. After a few residual blocks, the feature maps are obtained.
Upon the feature map, we add a region proposal network for object detection. The classifier is used to predict the object category of the bounding boxes while the box regressor is used to output the coordinates of the bounding boxes.

\subsection{Model Architecture}
Our model, DetTransNet, is implemented on top of the Vision Transformer backbone. 
It has a detection head that is similar to Faster RCNN \cite{Ren15}.
The detection head is used to produce bounding box regression and classification similar to the general CNNs based detection methods.
Our DetTransNet is able to not only learn representations of images and the objects, but also transfer the information learned to more tasks like object detection.
This is helpful for many computer vision models.

\begin{figure*}
    \centering
    \includegraphics[width=0.95\textwidth]{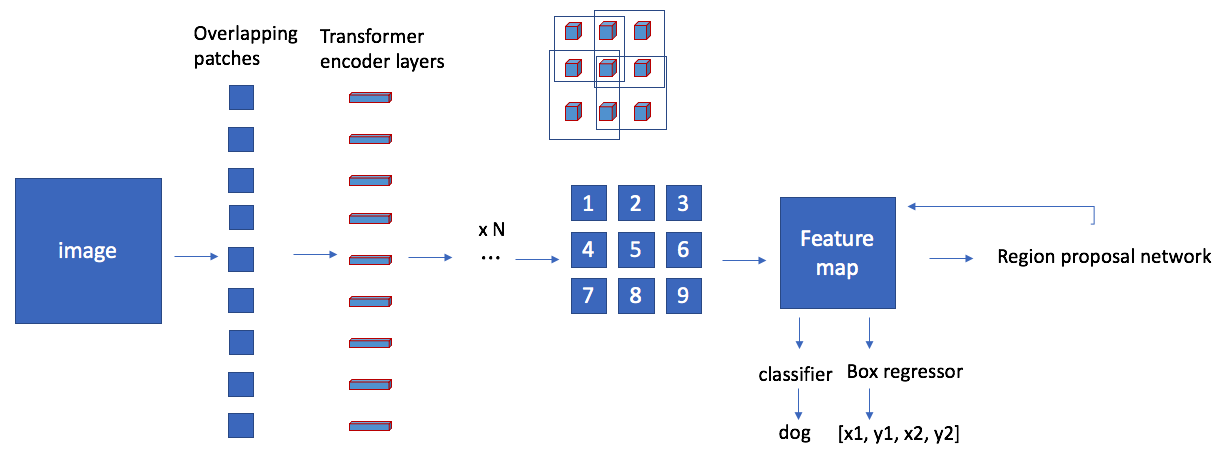}
    \caption{Model architecture. We use the Vision Transformer as the backbone and add a detection head for object detection. The DetTransNet is efficient and effective in many computer vision related tasks.}
    \label{fig:model}
\end{figure*}

In a general Visual Transformer, only the class token representation from the final transformer layer is used. All the other tokens are discarded because they are not related to the class prediction task.
Because the discarded tokens are related to the image patches from different locations, they might have encoded rich object and location information which is important for the object detection task.
Therefore, in order to fully utilize all the final output features, we propose to reorder them into a square feature map, which is later used as the input to a detection head similar to Faster RCNN.
We use a region proposal network to predict the objects in the image.
Some of the proposals are used as the input for a classification task.
Thus, our DetTransNet can precisely detect the objects as well as classify them into different categories.
The DetTransNet is shown in Figure \ref{fig:model}.

\subsection{Vision Transformer Backbone}
Vision Transformer \cite{dosovitskiy2021an} has proved that a transformer architecture can also be used in computer vision tasks besides the general natural language processing tasks. It achieves this by taking images as sequences of image patches. Even though the performance is not state-of-the-art yet, it opens a door to connect computer vision methods with natural language processing methods.

The Vision Transformer backbone is shown in Figure \ref{fig:vit}.

The vision transformer first reshapes the image of size $H \times W \times C$ into a sequence of flattened $2D$ image patches of shape $N \times P \times P \times C$. $H$ is the image height. $W$ is the image width. $C$ is the image channel. $P$ is the patch size and $N$ is the number of patches, which is also the input length.
Then a linear layer is used to flatten the image patches into vectors of dimension $D$.
Then a general transformer encoder is used to map the vectors to image representations $z$. 
Similar to BERT \cite{devlin-etal-2019-bert}, the vision transformer also used a learnable embedding to the sequence of the embedded patches, which corresponds to a classification token.
It also uses positional encoding to help the transformer remember the input sequence orders.
The Transformer encoder \cite{Vaswani2017,yang2019latent} consists of several multiheaded self-attention layers. Layer normalization is applied before
every block, and residual connections after every block.

Because the two dimensional neighborhood structure is used sparsely, and in order to increase the inductive bias of the transformer model, \cite{dosovitskiy2021an} explores how to increase the inductive bias and also different hybrid model architectures, where the input sequence can be formed
from feature maps of a CNN. A special case is that the patches can have spatial size $1 \times 1$, which means that the input sequence is obtained by simply flattening the spatial dimensions of the feature map and projecting to the Transformer dimension.
For more details of the model, please refer to \cite{dosovitskiy2021an}.

\subsection{Effective Patching Method}
The general vision transformer suffers from two weaknesses. Firstly, it requires much longer training time than conventional CNN based detectors. Secondly, it achieves low detection performance on small objects.
The weaknesses come from the patching and sequencing method that the images are divided into patches which does not contain useful location information.
In our method, we address this by introducing a new and effective patching method.
In short, we use overlapping patches. A single path shares $m$ pixels with its neighbor patches. By chaning the number $m$, it can share more information with the neighbors.
This can also help attend to small sampling locations in the feature maps.

\subsection{Region Proposal Network}
The objective of Region proposal network is to take an image as input and output object proposals. All of them have an objectness score to signify whether the bounding box contains an object.
Similar to Faster RCNN, we use fully convolutional network \cite{fully}.
Region proposal network slides a small kernel of size $s \times s$ over the feature maps of the residual network.
Each kernel is a sliding window whose output is fed into two fully connected layers, a box regressor and a box classifier.

\section{Experiments}
\label{sec:exp}
We run extensive experiments to evaluate the effectivenss of our proposed DetTransNet on the popular COCO 2017 detection dataset \cite{Lin14,reformer}.
We also analyze the performance of our model compared with other baselines to demonstarte the advantages of DetTransNet in both training efficiency and generalization ability.

\subsection{Datasets}
In this work, we evaluate our model and the compared baseline models on COCO 2017 object detection dataset. This dataset consists of over 5000 training and validation images and 5000 test images with over 20 object categories.
To pretrain our model with large datasets, we use ImageNet \cite{imagenet} and JFT dataset \cite{sun,xuewen_geo}.
ImageNet is a publicly released dataset consisting of more than 10 million images with labels, while JFT dataset is a larger dataset with over 300 million images. Using these two large datasets for pretraining can greatly improve the performance of our model as the small COCO dataset is way from enough.

\subsection{Experimental Settings}
We use the same training settings as the Mask R-CNN \cite{mask}. Images are resized
such that their shorter edge is 800 pixels.
We use 8 Nvidia V-100 GPUs with batch size of 2 images per GPU.
Each image has 64 sampled ROIs.
We train our model for 200 $k$ iterations with a learning rate of $0.001$. We use Adam optimizer \cite{kingma2015, xuewen_mm20} with a weight decay of $0.0001$ and momentum of $0.9$.
For training efficiency, region proposal network is first trained and then fixed for the rest of the training time.
For the hyper-parameters, we search for optimal hyperparameters through ablation studies.

\begin{table}[!t]
\footnotesize
\centering
\begin{tabular}{ccccccl}
\toprule
Method & AP & AP small & AP medium & AP large  \\
\midrule
R-CNN & 38.0 & 17.5 & 40.8 & 56.1 \\
Fast R-CNN & 39.8 & 18.4 & 42.3 & 58.7 \\
Faster R-CNN & 42.0 & 20.5 & 45.8 & 61.1  \\
YOLO & 41.9 & 19.6 & 45.6 & 60.8 \\
DETR & 42.0 & 20.5 & 45.8 & 61.1 \\
De-DETR & 43.8 & 21.2 & 46.5 & 61.9 \\
DetTransNet & \textbf{45.4} & \textbf{22.5} & \textbf{47.8} & \textbf{62.1} \\

\bottomrule
\end{tabular}
\caption{Experimental results on COCO object detection dataset. Best performance is in \textbf{bold}.}
\label{tab:coco}
\vspace{-0.1cm}
\end{table}

\subsection{Experimental Analysis}
We show our results in table \ref{tab:coco}. As we can see, our proposed DetTransNet outperformed all the other baselines, setting a new state-of-the-art results of $45.4$, $22.5$, $47.8$ and $62.1$. Compared with De-DETR \cite{zhu2021deformable}, we achieved $1.6$, $1.3$, $1.2$ and $0.2$.
These results demonstrated the effectiveness and efficiency of our model.
It also suggests that the overlapping image patches is a sound technique to solve the weaknesses of other vision transformer based object detectors.
CNN based models though efficient, are not effective as the vision transformer based object detectors.
\section{Conclusion}
\label{sec:con}
In this work, we presented a novel DetTransNet model, that is a vision transformer based object detector with rigion proposal network.
Our proposed overlapping patches technique is effective in solving the weaknesses of the general vision transformer detector.
Our future work includes how to train the model more efficiently and improves the FPS.

\bibliography{example_paper}
\bibliographystyle{icml2019}

\end{document}